\documentclass[runningheads]{llncs}
\usepackage[utf8]{inputenc}
\usepackage[T1]{fontenc}
\usepackage{graphicx}
\usepackage{amsmath}
\usepackage{graphicx}
\usepackage{booktabs}
\usepackage{amsmath}
\usepackage{soul}
\usepackage{amssymb}
\usepackage{multirow}
\usepackage{hyperref}
\newcommand{\header}[1]{#1}
\usepackage{natbib}
\usepackage{tikz}
\usepackage{enumitem} 
\usetikzlibrary{shadows} 
\usetikzlibrary{backgrounds} 
\usetikzlibrary{positioning}  

\usetikzlibrary{arrows.meta, positioning, shapes.geometric, calc}

\begin{document}

\title{From Symbolic to Neural and Back: 
Exploring Knowledge Graph–Large Language Model Synergies}

\titlerunning{Exploring KG–LLM Synergies}

\author{Bla\v{z} \v{S}krlj\inst{1}\orcidID{0000-0002-9916-8756} \and
Boshko Koloski\inst{1,2}\orcidID{0000-0002-7330-0579} 
\and \\ 
Senja Pollak\inst{1}\orcidID{0000-0002-4380-0863} 
\and Nada Lavra\v{c}\inst{1}\orcidID{0000-0002-9995-7093}}

\authorrunning{\v{S}krlj et al.}
\institute{(1) Jožef Stefan Institute \\ (2) Jožef Stefan International Postgraduate School \\
Jamova cesta 39, 1000 Ljubljana, Slovenia \\
\email{\{name.surname\}@ijs.si}}

\maketitle            

\begin{abstract}
Integrating structured knowledge from Knowledge Graphs (KGs) into Large Language Models (LLMs) enhances factual grounding and reasoning capabilities. This survey paper systematically examines the synergy between KGs and LLMs, categorizing existing approaches into two main groups: KG-enhanced LLMs, which improve reasoning, reduce hallucinations, and enable complex question answering; and LLM-augmented KGs, which facilitate KG construction, completion, and querying. Through comprehensive analysis, we identify critical gaps and highlight the mutual benefits of structured knowledge integration. Compared to existing surveys, our study uniquely emphasizes scalability, computational efficiency, and data quality. Finally, we propose future research directions, including neuro-symbolic integration, dynamic KG updating, data reliability, and ethical considerations, paving the way for intelligent systems capable of managing more complex real-world knowledge tasks.

\end{abstract}


\section{Introduction}

Knowledge Graphs and Large Language Models represent two foundational \textit{Artificial Intelligence} (AI) and \textit{Natural Language Processing} (NLP) paradigms. On the one hand, \textbf{Knowledge Graphs} (KGs) provide a structured and semantically rich representation of information by encoding entities as nodes and their interactions as edges. This graph-based representation enables computational systems to traverse complex information networks, facilitating improved data interpretation and contextualized reasoning. Such capabilities make KGs indispensable in domains that require a deep understanding of interconnected data, including semantic search, recommendation systems, drug design and comprehensive knowledge management~\cite{zeng2022toward,hogan2021knowledge,peng2023knowledge}.
On the other hand, \textbf{Large Language Models} (LLMs), exemplified by systems such as GPT-3~\cite{brown2020language}, employ advanced deep learning architectures to perform sophisticated linguistic tasks. These models are superior in natural language generation, comprehension and translation, relying on large-scale corpora and state-of-the-art neural architectures, most notably \textbf{Transformer networks}~\cite{vaswani2017attention}. Through the assimilation of large amounts of data, LLMs are capable of mastering syntax, semantics, and contextual dependencies, achieving unprecedented performance in tasks ranging from text summarization to conversational interfaces~\cite{zhao2023survey}.

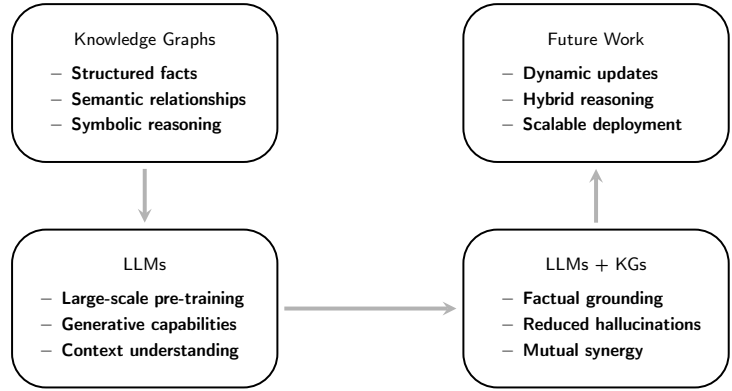
\begin{figure}[ht!]
    \centering
    \scalebox{0.75}{%
    \begin{tikzpicture}[
        >=stealth,
        font=\sffamily\footnotesize,
        every node/.style={
            rectangle,
            rounded corners=6mm,
            align=center,
            text width=4cm,
            minimum height=2.8cm,
            inner sep=10pt,
            line width=1pt
        },
        kgstyle/.style={
            fill=blue!0,
            draw=black!100,
            text=black!100
        },
        llmstyle/.style={
            fill=teal!0,
            draw=black!100,
            text=black!100
        },
        synergy/.style={
            fill=purple!0,
            draw=black!100,
            text=black!100
        },
        future/.style={
            fill=orange!0,
            draw=black!100,
            text=black!100
        },
        arrowstyle/.style={
            ->,
            line width=2pt,
            color=gray!60,
            shorten >=3pt,
            shorten <=3pt
        },
        header/.style={
            font=\sffamily\small\bfseries,
            color=black!100
        },
        edgelabel/.style={
            font=\sffamily\footnotesize\itshape,
            text=black!70,
            fill=white,
            inner sep=2pt
        }
    ]

    \node[kgstyle] (kg) at (-1,4) {%
        {\header Knowledge Graphs}\\[6pt]
        \begin{itemize}[leftmargin=20pt, itemsep=2pt, topsep=0pt]
            \item \textbf{Structured facts}
            \item \textbf{Semantic relationships}
            \item \textbf{Symbolic reasoning}
        \end{itemize}
    };

    \node[llmstyle] (llm) at (-1,0) {%
        {\header LLMs}\\[6pt]
        \begin{itemize}[leftmargin=15pt, itemsep=2pt, topsep=0pt]
            \item \textbf{Large-scale pre-training}
            \item \textbf{Generative capabilities}
            \item \textbf{Context understanding}
        \end{itemize}
    };

    \node[synergy] (llmkg) at (7,0) {%
        {\header LLMs + KGs}\\[6pt]
        \begin{itemize}[leftmargin=20pt, itemsep=2pt, topsep=0pt]
            \item \textbf{Factual grounding}
            \item \textbf{Reduced hallucinations}
            \item \textbf{Mutual synergy}
        \end{itemize}
    };

    \node[future] (future) at (7,4) {%
        {\header Future Work}\\[6pt]
        \begin{itemize}[leftmargin=20pt, itemsep=2pt, topsep=0pt]
            \item \textbf{Dynamic updates}
            \item \textbf{Hybrid reasoning}
            \item \textbf{Scalable deployment}
        \end{itemize}
    };

    \draw[arrowstyle] (kg.south) -- (llm.north);
    
    \draw[arrowstyle] (llm.east) -- (llmkg.west);
    
    \draw[arrowstyle] (llmkg.north) -- (future.south);
    \end{tikzpicture}
    }%
    \caption{Scope of this paper, illustrating the evolution from Knowledge Graphs to LLMs, their synergy, and directions for future work.}
    \label{fig:kg-llm-future}
\end{figure}

This work surveys the intersection of these two paradigms, structured Knowledge Graphs and foundational Language Models, and examines their complementary potentials. By integrating the structured and semantically coherent framework of KGs with the generative capabilities of LLMs, it becomes feasible to construct hybrid systems that are both contextually aware and computationally robust. On the one hand, the structured information encapsulated in KGs may serve as a guiding framework for LLMs, ensuring that generated language is accurate and semantically grounded. On the other hand, LLMs can help in the generation, augmentation, and maintenance of KGs by extracting pertinent information from unstructured textual sources and systematically organizing it into coherent graph structures.
The integration of these technologies can advance numerous applications: enhanced natural language understanding, improved techniques for information retrieval~\cite{fei2021enriching}, and sophisticated knowledge discovery methodologies~\cite{alam2022language} are just a few of the many promising use cases. The logical structure of KGs, when combined with the contextual reasoning of LLMs, paves the way for AI systems to achieve greater precision, richer semantic interactions, and advanced analytical capabilities.
In the following sections, summarized in Figure~\ref{fig:kg-llm-future}, this work systematically explores the construction and refinement of knowledge graphs, the underlying LLM architectures and the methodologies by which these two paradigms can be effectively combined, resulting in a conceptual framework and practical guiding principles for scholars and practitioners interested in exploiting the potential of KG and LLM integration. By undertaking this research, we seek to advance the field’s understanding of these two individual AI paradigms, their cross-fertilization, and their capacity to jointly expand the boundaries of AI research and applications.

\section{Knowledge Graphs}

Mining knowledge graphs involves extracting, processing, and organizing data from various sources to build a comprehensive and structured graph representation. Key techniques include \textbf{entity recognition}, \textbf{entity linking}, \textbf{relationship extraction}, and \textbf{data integration} \cite{Hogan2021KnowledgeGraphs,Zhao2024SurveyKGC}. 

\textbf{Entity recognition} involves identifying and classifying entities (e.g., people, organizations, locations) within text. This step is crucial for populating knowledge graphs with accurate and relevant data. Advanced techniques, such as named entity recognition (NER) models, leverage machine learning and natural language processing to achieve high accuracy in identifying entities \cite{Hogan2021KnowledgeGraphs,Sequeda2021EnterpriseKG}. These models can be trained on large annotated datasets to recognize various types of entities in different domains. As an illustration, Table~\ref{tbl:concepts} presents a random sample of triplets related to the concept ``mountain'', taken from the ConceptNet~\cite{speer2017conceptnet} knowledge graph. 

\begin{table}[b!]
\centering
\caption{A random sample of head-relation-tail triplets related to the concept ``mountain'' from the ConceptNet knowledge graph~\cite{speer2017conceptnet}.}
\resizebox{\textwidth}{!}{%
\begin{tabular}{@{}ccc@{}}
\toprule
\textbf{head} & \textbf{relation} & \textbf{tail} \\ \midrule
\texttt{/c/fr/eucalyptus\_camphora/n/wn/plant}   & \texttt{/r/Synonym}    & \texttt{/c/en/mountain\_swamp\_gum/n/wn/plant/} \\
\texttt{/c/en/mountain\_bike/n}                  & \texttt{/r/RelatedTo}  & \texttt{/c/en/traction/} \\
\texttt{/c/en/faith\_will\_move\_mountains}        & \texttt{/r/RelatedTo}  & \texttt{/c/en/god/} \\
\texttt{/c/es/diablo\_espinoso/n/wn/animal}       & \texttt{/r/Synonym}    & \texttt{/c/en/mountain\_devil/n/wn/animal/} \\
\texttt{/c/en/antiapennine/n}                     & \texttt{/r/RelatedTo}  & \texttt{/c/en/mountain\_range/} \\
\texttt{/c/en/climbing\_mountain}                & \texttt{/r/UsedFor}    & \texttt{/c/en/reaching\_top/} \\
\texttt{/c/pl/robić\_z\_igły\_widły/v}            & \texttt{/r/RelatedTo}  & \texttt{/c/en/make\_mountain\_out\_of\_molehill/} \\
\bottomrule
\end{tabular}
}
\label{tbl:concepts}
\end{table}

\textbf{Relationship extraction} focuses on identifying and categorizing the relationships between entities, such as “works at” or “located in”. This process is essential for building the connections within the graph that represent the interactions and associations between entities. Techniques for relationship extraction include rule-based methods, supervised learning algorithms, and deep learning models \cite{Dong2014KnowledgeVault}. These methods can analyze the context and syntactic structure of sentences to accurately determine the nature of relationships between entities.

\textbf{Entity linking} focuses on extracting entities from free-form text and correctly mapping them to the corresponding entries in a knowledge graph, such as Wikidata. For example, in the sentence "Jeff Bezos founded Amazon," entity linking maps "Amazon" to the company Amazon (Wikidata entity Q3884) and "Jeff Bezos" to the person Jeff Bezos (Wikidata entity Q312556). Entity linking methods vary from specialist models to generalist models that perform simultaneous relation extraction and entity linking, with state-of-the-art models being built on transformer architecture ~\cite{sevgili2022neural}.

\textbf{Data integration} combines information from multiple sources to create a unified knowledge graph. This step involves aligning and merging data from diverse datasets, resolving conflicts, and ensuring consistency. Data integration techniques include schema matching, entity resolution, and data fusion \cite{Hogan2021KnowledgeGraphs}. Schema matching aligns different data schemas to a common representation, while entity resolution identifies and merges duplicate entities across datasets. Data fusion combines information from multiple sources, selecting the most reliable data, and resolving inconsistencies.

Knowledge graphs are used in a wide range of applications, including semantic search, recommendation systems, and question-answering systems. Semantic search improves the accuracy of search engines by understanding the meaning behind queries. Unlike traditional keyword-based search, semantic search leverages relationships and context within knowledge graphs to deliver more relevant results \cite{Pound2010AdHocSearch}. For example, a query about “famous scientists in the 20th century” can return results that include notable figures and their contributions, even if the exact keywords are not present in the search query.
\textbf{Recommendation systems} use knowledge graphs to provide personalized suggestions based on user preferences and behaviors. By analyzing the relationships between users, items, and various attributes, knowledge graphs enable recommendation systems to offer more accurate and relevant recommendations \cite{Guo2022KGRecSurvey}. For instance, a movie recommendation system can suggest films based on a user's viewing history, preferences, and the relationships between movies, actors, and genres within the knowledge graph.
Question-answering systems leverage knowledge graphs to deliver precise and contextually relevant responses to user queries. These systems can traverse the graph to find the most accurate information, considering the relationships and connections between entities \cite{Diefenbach2018QA}. For example, a question about “the capital of France” can be answered by navigating the knowledge graph to find the entity “France” and its associated attribute “capital”.


To address the rigidity and sparsity of purely symbolic knowledge graphs, we turn to knowledge graph embeddings (KGEs), which map entities and relations into continuous vector spaces and enable efficient numerical reasoning for downstream machine learning tasks such as link prediction, entity classification, and clustering. Popularised by the representational learning paradigm~\cite{replearn}, KGEs facilitate the application of symbolic knowledge to tasks such as text classification~\cite{KOLOSKI2022208} and enrich representations for tabular data~\cite{carte}, enabling efficient interaction between structured semantic knowledge and diverse downstream modalities.

Early models such as TransE and its variants (TransH, TransR, TransD) treated relationships as translations in embedding space. More recent techniques, such as ConvE and RotatE, have improved on these by leveraging convolutional networks and rotational transformations to capture complex relational patterns \cite{Wang2021Survey,Ferrari2022Comprehensive}. 
Training strategies significantly impact KGE effectiveness. Ruffinelli et al. \cite{Ruffinelli2020OldDog} demonstrated that with proper hyperparameter tuning, even older models like RESCAL and TransE can match or outperform newer architectures. A major direction in embedding research is the incorporation of relation paths, as shown in PConvKB, which integrates path-based attention to improve embeddings \cite{Jia2020RelationPaths}. Another breakthrough, PairRE, introduced the concept of paired relation vectors to enhance the representation of N-to-N and 1-to-N relationships \cite{Chao2021PairRE}. 

Recent advances continue to refine KGE methods. Li and Zhu \cite{Li2024TransEMTP} proposed TransE-MTP, which builds on TransE but incorporates \textbf{multi-translation principles} to better capture hierarchical and compositional structures. The study of \textbf{benchmarking KGE models} has also gained traction, with systematic evaluations highlighting efficiency vs. effectiveness trade-offs across models \cite{Ferrari2022Comprehensive}. 
Mining knowledge graphs involves a combination of \textbf{entity recognition}, \textbf{relationship extraction}, and \textbf{data integration} techniques to build a structured representation of information. These knowledge graphs are invaluable in various applications, enhancing the accuracy and relevance of search, recommendation, and question-answering systems. 

However, KGs present several inherent challenges. Often building and maintaining their structure demands domain experts—a costly and often scarce resource—especially when handling temporal information, which complicates entity linking and versioning. Additionally, most efforts rely on a fixed ontology, making later revisions labor-intensive and error-prone. Real-world data scarcity and heterogeneity further hinder graph construction, as missing or inconsistent data require careful mapping and manual correction. Finally, merging graphs or expanding them across diverse subdomains frequently exposes mismatches in terminology, granularity, and temporal scope, undermining seamless integration.

\section{Large Language Models (LLMs)}
Large Language Models (LLMs) represent a significant breakthrough in natural language processing (NLP) and artificial intelligence (AI), enabling machines to understand, generate, and interact using human language with remarkable fluency. These models leverage deep learning architectures to process large amounts of textual data, learning intricate linguistic patterns, factual knowledge, and contextual relationships. One of the most notable advances in this domain has been the adoption of the \textbf{Transformer architecture}~\cite{vaswani2017attention}, which introduced self-attention mechanisms that allow models to capture long-range dependencies efficiently. 

Modern LLMs, such as OpenAI's GPT-3, which boasts 175 billion parameters \cite{brown2020gpt3}, have demonstrated unprecedented capabilities across a wide range of language-related tasks, including text completion, machine translation, summarization, question-answering, and even creative writing. These models undergo a two-step process: pre-training on a large corpus of diverse text followed by fine-tuning on specific tasks. The pre-training phase is often conducted using self-supervised learning objectives such as masked language modeling (MLM) in BERT \cite{devlin2019bert} or autoregressive token prediction in GPT \cite{radford2019gpt2}. The fine-tuning phase further refines the model using supervised datasets tailored to domain-specific tasks, significantly improving performance.
LLMs are referred to as \textbf{foundation models} \cite{bommasani2021foundation}, which means that they serve as a general-purpose base upon which various downstream applications can be built. The scalability of these models, following well-documented scaling laws \cite{brown2020gpt3}, suggests that increasing model size, dataset size, and compute resources leads to predictable improvements in language understanding and generation. However, these models also raise challenges concerning ethical AI, bias mitigation, interpretability, and computational efficiency, topics that continue to be actively researched.

To understand how LLMs achieve their capabilities, we explore their fundamental components, including attention mechanisms, token embeddings, and positional encodings.
Attention mechanisms have been a transformative innovation in NLP, allowing models to dynamically weigh the importance of different words in a given input sequence \cite{bahdanau2015nmt}. This idea was first introduced in the context of neural machine translation, where it enabled models to attend to relevant words in the source sentence while generating the target sentence. The Transformer architecture expanded upon this idea with the introduction of \textit{self-attention} \cite{vaswani2017attention}, where each word attends to every other word in the sequence, regardless of distance. 
The Transformer’s multi-head self-attention mechanism further enhances this capability by computing multiple attention scores in parallel, each capturing different linguistic relationships \cite{vaswani2017attention}. This is a fundamental departure from recurrent architectures, which struggle with long-range dependencies due to vanishing gradient issues. The ability to capture complex syntactic and semantic relationships has made the Transformer the de facto standard for modern NLP models.

Words and phrases in human language need to be represented in a numerical form that models can process. This is achieved through token embeddings, which map words, subwords, or characters into dense vector spaces. Early approaches to word embeddings, such as Word2Vec \cite{mikolov2013distributed} and GloVe \cite{pennington2014glove}, demonstrated that words with similar meanings often have similar vector representations. 
While early word embeddings were static (i.e. the representation of a word was fixed regardless of context), modern LLMs leverage \textbf{contextual embeddings}, where the meaning of a word adapts based on surrounding words. This advancement, introduced by models like BERT \cite{devlin2019bert}, allows disambiguation of homonyms and polysemous words. For example, in the sentence “The bank is located near the river”, the meaning of “bank” is different from its meaning in “I deposited money in the bank”. Contextual embeddings ensure that each instance of “bank” receives a different vector representation depending on its usage.
LLMs have revolutionized numerous NLP tasks, spanning both general and domain-specific applications. Some of the most significant use cases are listed below. 
The Transformer architecture has significantly advanced machine translation, outperforming earlier sequence-to-sequence recurrent models \cite{vaswani2017attention}. By leveraging self-attention, LLMs capture \textbf{long-range dependencies} and subtle linguistic nuances, enabling accurate translation across multiple languages. Google's Transformer-based translation models have set new benchmarks in translation quality, achieving human-like fluency in certain language pairs.
Text summarization is essential for condensing lengthy documents into concise versions while preserving key information. Models like BERTSUM \cite{devlin2019bert} and T5 \cite{raffel2020t5} have demonstrated remarkable proficiency in extractive and abstractive summarization. Extractive summarization selects the most relevant sentences from a document, while abstractive summarization generates new text that captures the original meaning concisely.
Sentiment analysis involves determining the emotional tone of text, which is widely used in social media monitoring, customer feedback analysis, and political discourse analysis. Transformer-based models fine-tuned on sentiment classification datasets, such as IMDb movie reviews or Twitter sentiment datasets, achieve state-of-the-art accuracy \cite{devlin2019bert}.

One of the most groundbreaking applications of LLMs is in conversational AI, where models like GPT-3 \cite{brown2020gpt3} and Google’s LaMDA \cite{thoppilan2022lamda} generate human-like dialogue. These models are used in chatbots, virtual assistants, and interactive agents, enhancing customer service, education, and entertainment experiences.
LLMs have also been tailored for domain-specific applications, such as scientific literature analysis and biomedical NLP. SciBERT \cite{beltagy2019scibert}, trained on scientific papers, excels at processing technical language, while BioBERT, specialized for biomedical texts, enhances tasks like named entity recognition and question answering in medicine. Recently, open source movement targeting accessible language models has given rise to many high-quality models. Selected examples are shown in Table~\ref{tbl:llmt}.

\begin{table}[t!]
\centering
\caption{Selected top 10 Open Source LLMs with Parameter Counts and Licenses.}
\resizebox{\textwidth}{!}{%
\begin{tabular}{l l l c c c}
\toprule
\textbf{Model} & \textbf{License} & \textbf{Developer} & \textbf{Parameter Count} & \textbf{Year} & \textbf{BibTeX Citation} \\
\midrule
LLaMA 3.1      & Community       & Meta AI              & 405B  & 2024 & \cite{meta2024llama3} \\
Mixtral 8x22B  & Apache 2.0      & Mistral AI           & 141B  & 2024 & \cite{mistral2024mixtral} \\
Command R      & Open Source (Research) & Cohere      & 35B   & 2024 & \cite{cohere2024commandr} \\
Command R+     & Open Source (Research) & Cohere      & 103B  & 2024 & \cite{cohere2024commandrplus} \\
Gemma          & Apache 2.0      & Google DeepMind      & 7B    & 2024 & \cite{deepmind2024gemma} \\
Jamba          & Apache 2.0      & AI21 Labs            & 52B   & 2024 & \cite{ai21labs2024jamba} \\
DBRX           & Apache 2.0      & Databricks Mosaic ML & 132B  & 2024 & \cite{databricks2024dbrx} \\
Phi-3 Medium   & MIT             & Microsoft            & 14B   & 2024 & \cite{microsoft2024phi3} \\
Nemotron-4     & Apache 2.0      & Nvidia               & 340B  & 2024 & \cite{nvidia2024nemotron4} \\
DeepSeek-V3    & MIT             & DeepSeek             & 671B  & 2024 & \cite{deepseek2024v3} \\
\bottomrule
\end{tabular}%
}
\label{tbl:llmt}
\end{table}

\section{KG-LLM Integration}

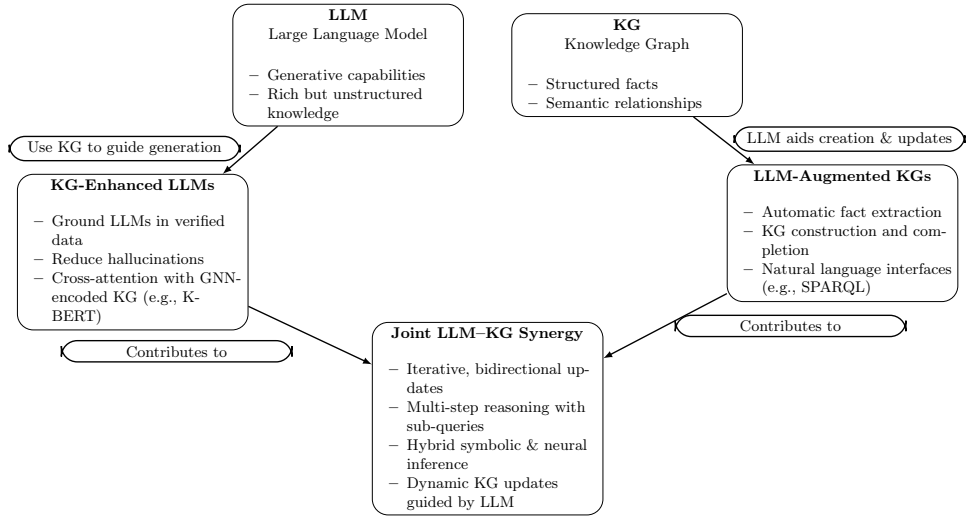
\begin{figure}[b!]
    \centering
    \scalebox{0.65}{%
    \begin{tikzpicture}[
        >=latex, 
        node distance=2em,
        every node/.style={
          draw,
          rectangle,
          rounded corners=1em,  
          align=center,
          text width=4.5cm
        },
        concept/.style={fill=white!15},
        synergy/.style={fill=white!25},
        base/.style={fill=white!35}
    ]

    \node[base] (llm) {%
      \textbf{LLM} \\
      \footnotesize Large Language Model \\
      \vspace{0.3em}
      \begin{itemize}
        \item Generative capabilities
        \item Rich but unstructured knowledge
      \end{itemize}
    };

    \node[base, right=3em of llm] (kg) {%
      \textbf{KG} \\
      \footnotesize Knowledge Graph \\
      \vspace{0.3em}
      \begin{itemize}
        \item Structured facts
        \item Semantic relationships
      \end{itemize}
    };

    \node[concept, below left=3em and -1em of llm] (kgenhanced) {%
      \textbf{KG-Enhanced LLMs} \\
      \footnotesize
      \begin{itemize}
        \item Ground LLMs in verified data
        \item Reduce hallucinations
        \item Cross-attention with GNN-encoded KG (e.g., K-BERT)
      \end{itemize}
    };

    \node[concept, below right=3em and -1em of kg] (llmaugmented) {%
      \textbf{LLM-Augmented KGs} \\
      \footnotesize
      \begin{itemize}
        \item Automatic fact extraction
        \item KG construction and completion
        \item Natural language interfaces (e.g., SPARQL)
      \end{itemize}
    };

    \node[synergy, below=5em of {$(kgenhanced)!0.5!(llmaugmented)$}] (joint) {%
      \textbf{Joint LLM--KG Synergy} \\
      \footnotesize
      \begin{itemize}
        \item Iterative, bidirectional updates
        \item Multi-step reasoning with sub-queries
        \item Hybrid symbolic \& neural inference
        \item Dynamic KG updates guided by LLM
      \end{itemize}
    };

    \draw[->, thick] (llm) -- node[left, xshift=-0.6em, yshift=0.1em]
      {\footnotesize Use KG to guide generation} (kgenhanced);

    \draw[->, thick] (kg) -- node[right, xshift=0.6em, yshift=0.1em]
      {\footnotesize LLM aids creation \& updates} (llmaugmented);

    \draw[->, thick] (kgenhanced) -- node[left, xshift=-1.1em, yshift=-1em]
      {\footnotesize Contributes to} (joint);

    \draw[->, thick] (llmaugmented) -- node[right, xshift=0.6em]
      {\footnotesize Contributes to} (joint);

    \end{tikzpicture}
    }%
    \caption{The interplay between Large Language Models (LLMs) and Knowledge Graphs (KGs).}
    \label{fig:llm-kg-concept-map}
\end{figure}

The intersection of LLMs and KGs has emerged as a promising area of research, aimed at combining the generalization power of LLMs with the precision and symbolic reasoning of KGs. The integration may strengthen the strengths, and weaken the weaknesses of the individual technologies:
\begin{itemize}
\item LLMs are at the forefront of AI-driven language understanding, shaping the future of human-computer interaction and knowledge discovery.
Large Language Models (LLMs) such as GPT-4 have achieved impressive results in generating human-like text and answering questions across domains. However, purely neural LLMs often act as black-boxes and may \textbf{hallucinate} incorrect facts or lack access to up-to-date knowledge. 
\item In contrast, a Knowledge Graph (KG) is a structured repository of facts represented as a graph of entities and relationships; KGs offer interpretability and reliable information, but are typically incomplete and costly to curate. 
\end{itemize}
\noindent By integrating LLM and KG technologies, we can build systems that answer questions and generate content with both the fluency of natural language and the correctness of evidence-backed knowledge. 
Multiple recent works have explored different aspects of KG-LLM integration. For example, Pan et al.~\cite{Pan2024Roadmap} present a roadmap for unifying LLMs and KGs, describing frameworks where KGs assist LLMs and vice versa. Another survey by Pan et al.~\cite{Pan2023Opportunities} discusses key opportunities and challenges when bridging these models. Hu et al.~\cite{Hu2023KnowledgeEnhanced} focus on techniques for injecting knowledge (often from KGs) into pre-trained language models, and Yang et al.~\cite{Yang2024FactAware} propose methods to enhance LLM factual consistency using KG-based information. Each of these studies covers a subset of the full landscape. Table~\ref{tab:surveyComparison} provides a comparative overview of the topics addressed by these surveys, pointing the interested reader to the topics of relevance. We consider as non-applicable the topics which either weren't yet relevant at the time of writing a paper, or intentionally focus on a different topic (it's apparent there is no coverage).
\begin{table}[t!]
\centering
\caption{Coverage of topics in recent LLM-and-KG surveys: Pan et al. (2024)~\cite{Pan2024Roadmap}, Pan et al. (2023)~\cite{Pan2023Opportunities}, Hu et al. (2023)~\cite{Hu2023KnowledgeEnhanced}, Yang et al. (2024)~\cite{Yang2024FactAware}. \checkmark = topic is covered; $\times$ = not covered; -- = out of scope/not applicable.}
\label{tab:surveyComparison}
\resizebox{\textwidth}{!}{%
\begin{tabular}{lcccc}
\toprule
\textbf{Topic} & \textbf{Pan et al. (2024)} & \textbf{Pan et al. (2023)} & \textbf{Hu et al. (2023)} & \textbf{Yang et al. (2024)} \\
\midrule
Entity Linking \& Alignment      & \checkmark & \checkmark & --         & $\times$   \\
Relation \& Attribute Extraction & \checkmark & \checkmark & --         & \checkmark \\
KG Embedding (Representation)    & \checkmark & \checkmark & $\times$   & $\times$   \\
KG Completion (Link Prediction)  & \checkmark & \checkmark & $\times$   & \checkmark \\
Graph-to-Text Generation         & \checkmark & $\times$   & $\times$   & $\times$   \\
KG Question Answering            & \checkmark & \checkmark & $\times$   & \checkmark \\
KG-Enhanced LLMs (Knowledge-Injected) & \checkmark & \checkmark & \checkmark & \checkmark \\
LLM-Augmented KGs (KG tasks)     & \checkmark & \checkmark & --         & --         \\
Synergized LLM+KG (Bidirectional)& \checkmark & \checkmark & $\times$   & \checkmark \\
Hallucination \& Factual Accuracy & \checkmark & \checkmark & $\times$   & \checkmark \\
\bottomrule
\end{tabular}%
}
\end{table}
Building on the above paradigms, we organize the overview of the paradigms at the intersection of KGs and LLMs into three primary categories, summarized in Figure~\ref{fig:llm-kg-concept-map}. First, \textbf{KG-enhanced LLMs} use knowledge graphs to improve the performance of language models. Second, \textbf{LLM-augmented KGs} leverage language models to construct, enrich, or utilize knowledge graphs. Third, \textbf{joint LLM--KG frameworks} involve a deeper bidirectional coupling of LLMs and KGs. We next describe each of these categories, along with representative use cases and techniques.

\subsection{KG-Enhanced LLMs}

KG-enhanced LLMs represent an emerging research paradigm that seeks to bridge the gap between vast but sometimes imprecise knowledge stored in large language models (LLMs) and the structured curated information contained in knowledge graphs (KGs). In this approach, a KG acts as an external reliable repository of factual data that the LLM can query to enrich its responses or guide its reasoning process. For example, by retrieving relevant subgraphs from a KG such as Wikidata, DBpedia, or ConceptNet, an LLM-based question-answering system can incorporate precise facts (such as the familial relationships of a historical figure) into its generated output, to enhance factual consistency and reduce hallucinations \cite{Yang2024FactAware}.
Several integration strategies have been explored. A common method is to augment the input prompt with additional context extracted from a KG, to effectively ground the LLM’s generation in verified data. Alternatively, more sophisticated techniques embed KG-derived representations directly into the model’s latent space. In these approaches, adapter modules or fine-tuning procedures are used to inject structured knowledge into the LLM’s parameters, helping the model internalize and recall factual information more reliably \cite{Hu2023KnowledgeEnhanced}. 

Recent work has also explored the use of graph neural networks (GNNs) to encode retrieved KG subgraphs, which are then integrated with LLMs via cross-attention mechanisms \cite{Zhang2024KGReasoning}. Such techniques enable a form of “knowledge-guided reasoning”, where intermediate model representations are directly influenced by structured data from the KG. Early models like K-BERT \cite{liu2020kbert} and KnowBERT \cite{peters2019knowbert} (originally developed to enhance language representations) have inspired further innovations in this direction. Moreover, models such as ERNIE \cite{sun2019ernie} have demonstrated that pre-training on knowledge-enhanced corpora can yield significant improvements in downstream tasks, particularly in question-answering and logical inference.
Beyond language-only models, ~\cite{rita} recently demonstrated that KGs can be introduced to Vision-Language Models (VLMs) as a fallback mechanism when a VLM is uncertain and can be used as a knowledge base to further improve visual question answering.

\textbf{KG-enhanced LLMs} combine the generative versatility of LLMs with the precision of structured knowledge sources. This fusion not only mitigates issues like hallucinations but also improves the interpretability of outputs by offering traceable reasoning paths through the underlying KG. Another recent attempt that exploits KG-based prompting to improve language models that operate via chain of thought (CoT) is the MindMap project~\cite{wen2024mindmap}. This project showcases structured KG prompting techniques that improve LLM reasoning accuracy and reduce hallucinations.
As research in this area continues to mature, we expect further innovations that refine the synergy between unstructured language models and structured knowledge graphs.

\subsection{LLM-Augmented KGs}
In the opposite direction, researchers have employed LLMs to assist in creating, populating, and querying knowledge graphs. LLMs can serve as powerful information extractors: given large corpora of text, an LLM can be prompted or fine-tuned to identify entities and relationships, essentially converting unstructured text into structured triples for \textbf{KG construction}~\cite{chen-etal-2024-sac,petroni-etal-2019-language}. This approach has been used to rapidly expand KGs by mining textual resources for new facts, significantly reducing the manual effort typically needed for KG curation. However, one needs to be considered when using LLMs to populate KGs, as LLM-augmented KGs run the risk of propagating the model’s hallucinations and biases into the KG. Thus, rigorous validation and human-in-the-loop curation are essential to ensure the integrity and reliability of these automatically generated knowledge graphs.
 Furthermore, \textbf{KG completion} tasks (predicting missing links or attributes in an existing graph), LLMs can leverage their broad world knowledge to suggest likely connections between entities \cite{Pan2024Roadmap}. Another application is using LLMs to translate natural language queries or commands into formal queries that a KG can execute (e.g., converting a question into SPARQL query language). By acting as an intelligent interface, the LLM makes the KG accessible to users who do not know the KG's query language, enabling conversational \textbf{question answering} over the graph. Additionally, LLMs have been utilized for \textbf{KG-to-text generation} to produce coherent descriptions or summaries of specific entities or subgraphs. Such generated text can help end users understand KG data in natural language and can also serve as supplementary training data for language models.
Recent research has further extended the capabilities of LLM-augmented KGs by integrating explicit structural cues into the language models. For example, Zhang et al. \cite{Zhang2023KoPA} proposed a novel method where structural embeddings of entities and relations are injected into the LLM via a knowledge prefix adapter. This technique, named \textbf{structure-aware reasoning}, enables the LLM to combine textual context with the inherent graph structure, enhancing its ability to more accurately predict missing links in knowledge graph completion. By merging cross-modal structural information with raw text, the approach effectively mitigates issues such as hallucinations and improves the reliability of KG completion.
Another study by Zhu et al. \cite{Zhu2023LLMsKG} offers a comprehensive evaluation of LLMs for KG construction and reasoning. Their work demonstrates that while models like GPT-4 excel in general language understanding, their performance in structured tasks such as link prediction and multi-hop query answering can be significantly boosted through domain-specific prompt engineering and targeted fine-tuning. The study shows that LLMs can not only extract entities and relations but also perform complex reasoning over these structures, thereby supporting applications that require in-depth inference and dynamic knowledge retrieval.
\par The overviewed work indicates that LLMs can be used as external domain evaluators, inspecting the quality of knowledge graphs extracted from raw, unstructured text, performing on par with or superseding human performance~\cite{huang2025llmsgoodgraphjudger,TSANEVA2025104145}.
These advances illustrate that the synergy between LLMs and knowledge graphs is evolving beyond simple extraction and query translation. By incorporating structural embeddings and reasoning strategies, LLM-augmented KGs are becoming more robust and interactive. 

\section{Joint LLM--KG Synergy}
Large Language Models (LLMs) and Knowledge Graphs (KGs) have complementary strengths, motivating frameworks that tightly integrate the two for bidirectional reasoning~\cite{Pan2023Roadmap}. Recent works propose joint architectures where an LLM's neural reasoning is coupled with a KG's symbolic structure in multi-step, interactive loops. For example, Liu et al. present a dual reasoning paradigm pairing an LLM with a graph neural network (GNN) to perform collaborative multi-hop reasoning~\cite{Liu2025DualR}. In their framework, the GNN derives explicit relational chains from the KG, providing interpretable paths subsequently converted into a knowledge-enhanced prompt guiding the LLM’s inference~\cite{Liu2025DualR}. This division of labor---with symbolic traversal handled by the KG and nuanced language understanding by the LLM---achieves state-of-the-art results in knowledge-intensive question answering (QA), demonstrating how mutual reinforcement can curb hallucinations and improve correctness.

Other approaches grant LLMs direct capabilities to query and update KGs \textbf{during reasoning}. Markowitz et al. introduce Tree-of-Traversals, a zero-shot \textit{neuro-symbolic planning} algorithm equipping a black-box LLM with discrete actions to interface with a KG~\cite{Markowitz2024TreeTraversals}. At each step, the LLM decides either to \textit{expand a node} via a relation in the KG or to \textit{propose an answer}, dynamically constructing a reasoning tree. A search procedure evaluates and revisits these actions, ensuring high-confidence reasoning paths~\cite{Markowitz2024TreeTraversals}. This tight loop allows KG information to influence the LLM’s reasoning dynamically and vice versa, significantly improving multi-hop QA performance without fine-tuning.
Similarly, Li et al. propose Decoding-on-Graphs (DoG), constraining the LLM's reasoning chain to conform to actual KG paths~\cite{Li2024DoG}. Applying graph-derived constraints during decoding ensures that each reasoning step is grounded in factual KG context, yielding faithful and sound multi-step reasoning that is both interpretable and effective for KG-based QA. 

Another research direction employs KGs as navigation guides for LLM retrieval and reasoning. Ma et al. introduce Think-on-Graph 2.0, a retrieval-augmented generation framework aligning queries with KG structure to direct information retrieval~\cite{Ma2023ToG2}. Here, the KG functions as a map, helping the LLM select entities and relations for detailed textual evidence retrieval. This KG-guided retrieval enhances logical consistency and reduces errors in multi-hop queries, highlighting the efficacy of hybrid structured–neural systems. Additionally, KG-derived planning has been harnessed to enhance LLM step-by-step reasoning. Wang et al. devise a method to train LLMs with planning data extracted from KGs, improving their iterative retrieval and reasoning capabilities through targeted fine-tuning~\cite{Wang2024Planning}.
Emerging studies also explore mutual refinement, allowing continuous reciprocal updates between LLMs and KGs. Zhang et al. introduce Chain-of-Knowledge, integrating symbolic inference into LLMs by training on logical rules mined from KGs~\cite{Zhang2024ChainOfKnowledge}. This iterative symbolic–neural co-training strengthens logical consistency within LLM reasoning processes, yielding superior performance on reasoning benchmarks.

Cutting-edge research is moving towards tight integration of LLMs and KGs, from pipeline architectures injecting KG-derived facts into LLM prompts, interactive agents interweaving language generation and graph traversal, to joint training techniques aligning neural representations with symbolic logic. This joint synergy supports dynamic bidirectional interaction---KGs grounding LLM reasoning steps, and LLMs populating or navigating KGs---resulting in robust, explainable, and knowledgeable AI systems~\cite{Pan2023Roadmap,Markowitz2024TreeTraversals}.

\section{Interesting Open Problems for Future Research}

Based on the reviewed work, we have identified a list of open problems in the field of combining LLMs and KGs.

\begin{enumerate}

\item \textbf{Structured-Unstructured Alignment:} How can we seamlessly align structured information from knowledge graphs with the unstructured, free-form text generated by LLMs to enable robust joint reasoning?
\item \textbf{Neuro-Symbolic Integration:} How can we optimally fuse neural representations with symbolic reasoning, combining the strengths of both paradigms for more effective problem solving?
\item \textbf{Dynamic Knowledge Updating:} How can joint LLM–KG frameworks support continuous, automated updates of the knowledge graph, incorporating new information and corrections over time?
\item \textbf{Data Quality and Consistency:} What techniques can ensure that automatically extracted or augmented knowledge maintains high accuracy and consistency across diverse and evolving data sources?
\item \textbf{Hallucination Mitigation:} What novel approaches can leverage KG constraints to reduce LLM hallucinations and improve factual grounding in generated content?
\item \textbf{Scalability and Efficiency:} How can we design integrated systems that effectively scale both large language models and massive knowledge graphs, while managing computational and memory overhead?
\item \textbf{Interpretability and Explainability:} What methods can improve the interpretability of hybrid systems so that reasoning paths, which span both symbolic (KG) and neural (LLM) components, become transparent and traceable?

\end{enumerate}

\section{Final remarks}
We explore the integration of knowledge graphs (KGs) and large language models (LLMs), reviewing core methods such as relation extraction in KGs and self-attention architectures in LLMs. Our analysis highlights that combining structured KGs with context-aware LLMs enhances factual accuracy and consistency in language generation, while LLM-augmented KGs automate knowledge graph construction and maintenance.
A comparative analysis with existing surveys identifies coverage gaps and outlines key areas needing further investigation.
Despite encouraging results, challenges persist, including scalability, computational efficiency, continuous data quality, and ethical concerns like model bias. Future research should aim to optimize KG–LLM integration, advance neuro-symbolic methods, refine automated knowledge extraction, and establish standardized evaluation benchmarks.

\section*{Acknowledgments}
We acknowledge the financial support of the Slovenian Research and Innovation Agency (ARIS) through grants GC-0002 (Large Language Models for Digital Humanities), GC-0001 (Artificial Intelligence for Science) and the core research programme P2-0103 (Knowledge Technologies). The work of B.K. was supported by the Young Researcher Grant PR-12394. We thank the anonymous reviewers for the constructive feedback. This work was partially funded by ARIS project with reference number J4-4555.

\bibliographystyle{plain}
\bibliography{references}

\end{document}